\documentclass[runningheads]{llncs}
\usepackage{xcolor}
\usepackage{graphicx}
\usepackage{bm}
\newcommand{\vect}[1]{\boldsymbol{\mathbf{#1}}}
\usepackage{caption}
\usepackage{amsmath,amssymb}
\usepackage{tabularray}
\UseTblrLibrary{booktabs}
\usepackage{multirow}
\usepackage{marvosym}
\usepackage{color, colortbl}

\usepackage{hyperref}

\begin{document}

\title{On the use of Mahalanobis distance for out-of-distribution detection with neural networks for medical imaging}
\titlerunning{On Mahalanobis distance for OOD detection}
%
\author{Harry Anthony$^{[0009-0004-1252-7448],\textrm{1,(\Letter)} }$  \and Konstantinos Kamnitsas$^{1,2,3}$} 
\authorrunning{H. Anthony et al.}
%
\institute{{$^1$Department of Engineering Science, University of Oxford, Oxford, UK} \email{harry.anthony@eng.ox.ac.uk}  {\\ $^2$Department of Computing, Imperial College London, London, UK \\ $^3$School of Computer Science, University of Birmingham, Birmingham, UK } }%
\maketitle              
\begin{abstract}

Implementing neural networks for clinical use in medical applications necessitates the ability for the network to detect when input data differs significantly from the training data, with the aim of preventing unreliable predictions. The community has developed several methods for out-of-distribution (OOD) detection, within which distance-based approaches - such as Mahalanobis distance - have shown potential. This paper challenges the prevailing community understanding that there is an optimal layer, or combination of layers, of a neural network for applying Mahalanobis distance for detection of any OOD pattern. Using synthetic artefacts to emulate OOD patterns, this paper shows the optimum layer to apply Mahalanobis distance changes with the type of OOD pattern, showing there is no one-fits-all solution. This paper also shows that separating this OOD detector into multiple detectors at different depths of the network can enhance the robustness for detecting different OOD patterns. These insights were validated on real-world OOD tasks, training models on CheXpert chest X-rays with no support devices, then using scans with unseen pacemakers (we manually labelled 50\% of CheXpert for this research) and unseen sex as OOD cases. The results inform best-practices for the use of Mahalanobis distance for OOD detection.  
The manually annotated pacemaker labels and the project’s code are available at: \url{https://github.com/HarryAnthony/Mahalanobis-OOD-detection}

\end{abstract}
\section{Introduction}

Neural networks have achieved state-of-the-art performance in various medical image analysis tasks. Yet their generalisation on data not represented by the training data - out-of-distribution (OOD) - is unreliable \cite{hu_challenges_2020,perone_unsupervised_2019,zech_variable_2018}. In the medical imaging field, this can have severe consequences.
Research in the field of OOD detection \cite{ruff_unifying_2021} seeks to develop methods that identify if an input is OOD, acting as a safeguard that informs the human user before a potentially failed model prediction affects down-stream tasks, such as clinical decision-making - facilitating safer application of neural networks for high-risk applications.

One category of OOD detection methods use an \textbf{external model for OOD detection}. These include using \textit{reconstruction models} \cite{baur_autoencoders_2021,graham_denoising_2023,pawlowski_unsupervised_2018,pinaya_unsupervised_2022,schlegl_f-anogan_2019}, which are trained on in-distribution (ID) data and assume high reconstruction loss when reconstructing OOD data. Some approaches employ a \textit{classifier} to learn a decision boundary between ID and OOD data \cite{ruff_unifying_2021}. The boundary can be learned in an unsupervised manner, or supervised with exposure to pre-collected OOD data \cite{hendrycks_deep_2018,guha_roy_does_2022,steinbuss_generating_2021,tan_detecting_2022}. 
Other methods use \textit{probabilistic models} \cite{kobyzev_normalizing_2021} to model the distribution of the training data, and aim to assign low probability to OOD inputs.

Another category are \textbf{confidence-based methods} that enable discriminative models trained for a specific task, such as classification, to estimate uncertainty in their prediction. Some methods use the network's softmax distribution, such as MCP \cite{hendrycks_baseline_2018}, MCDropout \cite{gal_dropout_2016} and ODIN \cite{liang_enhancing_2020}, whereas others use the distance of the input to training data in the model's latent space \cite{lee_simple_2018}.

A commonly studied method of the latter category is Mahalanobis distance \cite{lee_simple_2018}, possibly due to its intuitive nature. The method has shown mixed performance in literature, performing well in certain studies \cite{gonzalez_distance-based_2022,kamoi_why_2020,rippel_gaussian_2021,uwimana1_out_2021} but less well in others \cite{berger_confidence-based_2021,song_rankfeat_2022,sun_dice_2022}. Previous work has explored which layer of a network gives an embedding optimal for OOD detection \cite{calli_frodo_2019,lee_simple_2018}. But further research is needed to understand the factors influencing its performance to achieve reliable application of this method. This paper provides several contributions towards this end:
\begin{itemize}
\item{Identifies that measuring Mahalanobis distance at the last hidden layer of a neural network, as commonly done in literature, can be sub-optimal.}
\item{Demonstrates that different OOD patterns are best detectable at different depths of a network, implying that there is no single layer to measure Mahalanobis distance for optimal detection of \emph{all} OOD patterns.}
\item{The above suggests that optimal design of OOD detection systems may require multiple detectors, at different layers, to detect different OOD patterns. We provide evidence that such an approach can lead to improvements.}
\item{Created a benchmark for OOD detection by manually annotating pacemakers and support devices in CheXpert \cite{irvin_chexpert_2019}.}

\end{itemize}

\section{Methods}
\label{sec:exp_meth}

\textbf{Primer on Mahalanobis score, $\mathcal{D}_{ \mathcal{M}}$:} Feature extractor $\mathcal{F}$ transforms input $\mathbf{x}$ into an embedding. $\mathcal{F}$ is typically a section of a  neural network pre-trained for a task of interest, such as disease classification, from which feature maps $h(\textbf{x})$ are obtained. The mean of feature maps $h(\textbf{x})$ are used as embedding vector $\vect{z}$: 
\begin{equation} \label{Feature_extractor}
   \vect{z} \in \Re^M = \frac{1}{D^2} \sum_{D} \sum_{D} h(\vect{x}), \ \ \ \textrm{where} \ h(\vect{x}) \in \Re^{D \times D \times M} 
\end{equation}
for $M$ feature maps with dimensions $D\!\times\!D$. Distance-based OOD methods assume the embedded in-distribution (ID) and OOD data will deviate in latent space, ergo being separable via a distance metric. In the latent space $\Re^M$, $N_c$ \emph{training} data points for class c have a mean and covariance matrix of

\begin{equation}
\label{eq:mu_sigma}
\vect{\mu_c} = \frac{1}{N_c} \sum_{i=1}^{N_c} \mathbf{z_{i_c}} , \ \ \vect{\Sigma_c} = \frac{1}{N_c} \sum^{N_c}_{i=1} (\mathbf{z_{i_c}} - \vect{\mu_{c}}) \ (\mathbf{z_{i_c}} - \vect{\mu_{c}})^T
\end{equation}
where $\vect{\mu_c}$ is vector of length $M$ and $\vect{\Sigma_c}$ is a $M\!\times\!M$ matrix. Mahalanobis distance $\mathcal{D}_{\mathcal{M}_c}$ between embedding $\mathbf{z}$ of a \emph{test} data point and the \emph{training} data of class c can be calculated as a sum over $M$ dimensions \cite{mahalanobis_generalised_1936}. The \textbf{Mahalanobis score} $\mathcal{D}_{\mathcal{M}}$ for OOD detection is defined as the minimum Mahalanobis distance between the test data point and the class centroids of the training data,
\begin{equation}
\label{eq:mah_dist}
\mathcal{D}_{\mathcal{M}_c}(\mathbf{x}) = \sum_{i=1}^M (\mathbf{z} - \vect{\mu_c}) \ \vect{\Sigma_c}^{-1} \ (\mathbf{z} - \vect{\mu_c})^{T}, \ \ \ \ \ \mathcal{D}_{\mathcal{M}}(\mathbf{x}) = \min_c { \{ \mathcal{D}_{ \mathcal{M}_c}(\mathbf{x}) \}}.
\end{equation}

\begin{figure}[h]
\centering
\includegraphics[width=\textwidth]{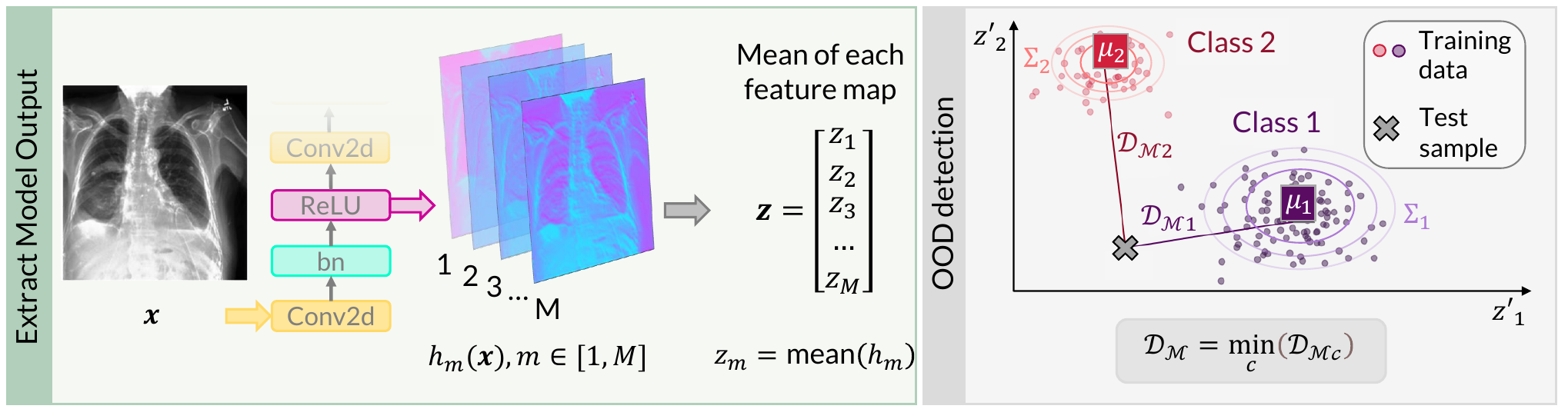}
\caption{(Left) Method to extract embeddings after a network module. (Right) Mahalanobis score $\mathcal{D}_{ \mathcal{M}}$ of an input to the closest training class centroid.
} \label{workflow_box}
\end{figure}

Threshold $t$, chosen empirically, is then used to separate ID ($\mathcal{D}_{\mathcal{M}}\!<\!t)$ from OOD data ($\mathcal{D}_{\mathcal{M}}\!>\!t$). Score $\mathcal{D}_{\mathcal{M}}$ is commonly measured at a network's last hidden layer (LHL) \cite{berger_confidence-based_2021,du_unknown-aware_2022,fort_exploring_2021,ren_simple_2021,song_rankfeat_2022,sun_dice_2022}.
To analyse the score's effectiveness with respect to where it is measured, we extracted a separate vector $\mathbf{z}$ after each network module (Fig. \ref{workflow_box}). Herein, a \emph{module} refers to a network operation: convolution, batch normalisation (BN), ReLU, addition of residual connections, pooling, flatten. Stats $\vect{\mu_c}^\ell$ and $\vect{\Sigma_c}^\ell$ (Eq.~\ref{eq:mu_sigma}) of the training data were measured after each module $\ell$, and for each input an OOD score $\mathcal{D}_{\mathcal{M}}^\ell$ was calculated per module (Eq.~\ref{eq:mah_dist}).

\textbf{Weighted combination:} Weighted combination of Mahalanobis scores $\mathcal{D}_{\mathcal{M}}^\ell$, measured at different layers $\ell$, was developed \cite{lee_simple_2018} to improve OOD detection:

\begin{equation}\label{coef}
    \mathcal{D}_{\mathcal{M}, \textrm{\textit{comb}}}(\mathbf{x}) = \sum_\ell \alpha_\ell \ \mathcal{D}_{\mathcal{M}}^{\ell}(\mathbf{x}),
\end{equation}
using $\alpha_l\! \in\! \Re$ to down-weight ineffective layers. Coefficients $\alpha_l$  are optimised using a logistic regression estimator on pre-collected OOD data \cite{lee_simple_2018}.

\textbf{Fast gradient sign method (FGSM) \cite{goodfellow_explaining_2015,lee_simple_2018}:} Empirical evidence showed that the rate of change of Mahalanobis distance with respect to a small input perturbation is typically greater for ID than OOD inputs \cite{lee_simple_2018,liang_enhancing_2020}. Therefore, perturbations $\vect{x'} = \vect{x} - \varepsilon \cdot \textrm{sign} (\nabla_x \mathcal{D}_{\mathcal{M}} (\vect{x} ))$ of magnitude $\varepsilon$ are added to image $\textbf{x}$, to minimise distance $\mathcal{D}_{\mathcal{M}_c}$ to the nearest class centroid. Mahalanobis score $\mathcal{D}_{\mathcal{M}}(\vect{x'})$ of the perturbed image $\vect{x'}$ is then used for OOD detection.

\textbf{Multi-branch Mahalanobis (MBM):} During this work it was found that different OOD patterns are better detected at different depths of a network (Sec.~\ref{sec:synth_results}).
This motivated the design of a system with multiple OOD detectors, operating at different network depths.
\
We divide a network into parts, separated by downsampling operations. We refer to each part as a \emph{branch} hereafter.
\
For each branch $b$, we combine (with summation) the scores $\mathcal{D}_{\mathcal{M}}^{\ell}$, measured at modules $\ell \! \in \! L_b$, where $L_b$ is the set of modules in the branch (visual example in Fig.~\ref{pacemaker_sex_AUROC}).
\
For each branch, we normalise each score $\mathcal{D}_{\mathcal{M}}^{\ell}$ before summing them, to prevent any single layer dominating. For this, the mean ($\mu_b^\ell=\mathbb{E}_{\mathbf{x}\in X_{train}}[D_M^\ell(\mathbf{x})]$) and standard deviation ($\sigma_b^\ell=\mathbb{E}_{x\in X_{train}}[(D_M^\ell(\mathbf{x})-\mu_b^\ell)^2]^{\frac{1}{2}}$) of Mahalanobis scores of the \emph{training data} after each module were calculated, and used to normalise $\mathcal{D}_{\mathcal{M}}^{\ell}$ for any \emph{test} image $\mathbf{x}$, as per Eq.~\ref{standardisation}.
This leads to a different Mahalanobis score and OOD detector per branch (4 in experiments with ResNet18 and VGG16).
\: 
\begin{equation}\label{standardisation}
    \mathcal{D}_{\mathcal{M}, \textrm{\textit{branch-b}}}(\mathbf{x}) = \sum_{\ell \in L_b} \frac{\mathcal{D}_{\mathcal{M}}^\ell(\mathbf{x}) - \mu_b^\ell}{\sigma_b^\ell}.
\end{equation}

\section{Investigation of the use of $\mathcal{D}_{ \mathcal{M}}$ with Synthetic Artefacts}
\label{sec:synth_results}

\begin{figure}
\centering
\begin{tabular}{cc}
 \includegraphics[width=3.5cm,height=3.1cm]{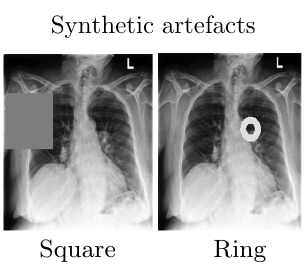}     \includegraphics[width=8.5cm,height=3.1cm]{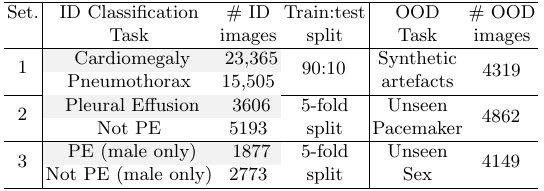} \\
 \hspace{-8em} a)  \hspace{17em}  b)
\end{tabular}
\caption{
a) Visual and b) quantitative summary of the synthetic (setting 1) and real (setting 2 \& 3) ID and OOD data used to evaluate OOD detection performance. 
}
\label{data_summary}
\end{figure}

The abilities of Mahalanobis score $\mathcal{D}_{ \mathcal{M}}$ were studied using CheXpert \cite{irvin_chexpert_2019}, a multi-label collection of chest X-rays. Subsequent experiments were performed under three settings, summarised in Fig. \ref{data_summary}. In the first setting, studied here, we used scans containing either Cardiomegaly or Pneumothorax. We trained a ResNet18 on 90\% of these images to classify between the two classes (ID task), and held-out 10\% of the data as ID test cases. We generated an OOD test set by adding a synthetic artefact at a random position to these held-out images.

\begin{figure}
\centering
 \includegraphics[width=\textwidth,height=3.625cm]{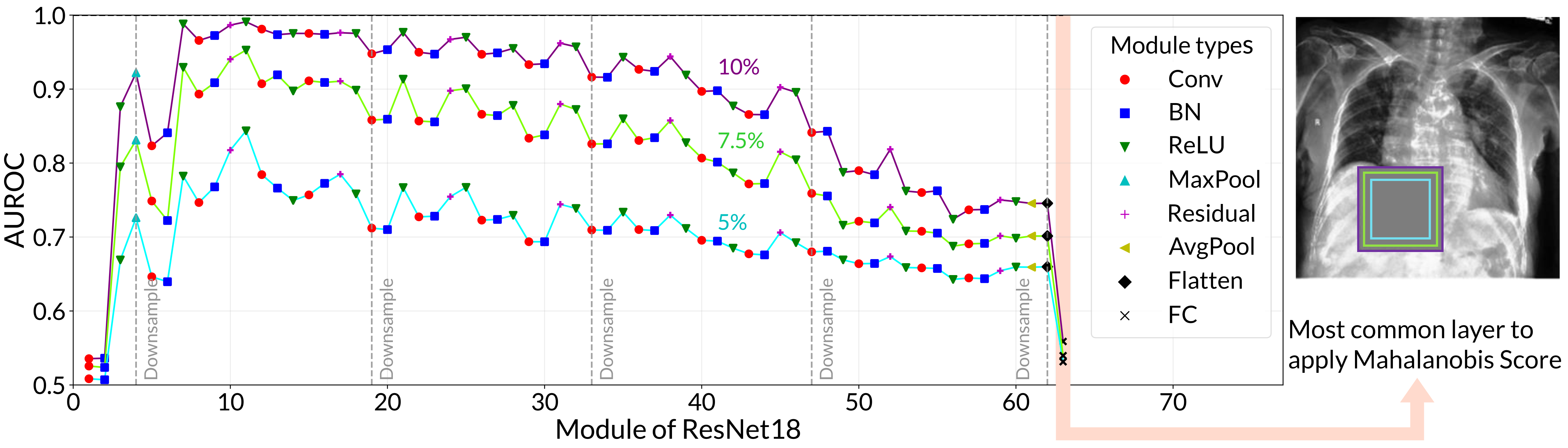} 
\caption{AUROC (mean of 3 seeds) for Mahalanobis score over the modules of ResNet18 for synthetic square artefacts of size 10\% (purple), 7.5\% (green) and 5\% (blue) of the image. The module types of ResNet18 are visualised, showing AUROC is typically improved after a ReLU module. The downsample operations are shown by dashed grey lines. The AUROC at the last hidden layer (LHL) is highlighted in orange, exhibiting a comparatively poor performance. }
\label{square_AUROC}
\end{figure}

\textbf{Square artefact:} Firstly, grey squares, of sizes 10, 7.5 and 5 \% of the image area, were introduced to create the OOD cases. We processed ID and OOD data, measured their $\mathcal{D}_{ \mathcal{M}}$ after every module in the network and plotted the AUROC score in Fig. \ref{square_AUROC}. We emphasize the following observations. The figure shows that larger square artefacts are easier to detect, with this OOD pattern being easier to detect in earlier layers. Moreover, we observed that AUROC is poor at the last hidden layer (LHL), which is a common layer to apply $\mathcal{D}_{ \mathcal{M}}$ in the literature \cite{berger_confidence-based_2021,du_unknown-aware_2022,fort_exploring_2021,ren_simple_2021,song_rankfeat_2022,sun_dice_2022}. The performance of this sub-optimal configuration may be diverting the community's attention, missing the method's true potential. The results also show AUROC performance in general improves after a ReLU module, compared to the previous convolution and BN of the corresponding layer. Similar results were found with VGG16 but not shown due to space constraints.

\begin{figure}
\centering
\begin{tabular}{cc}
 \includegraphics[width=\textwidth,height=3.8cm]{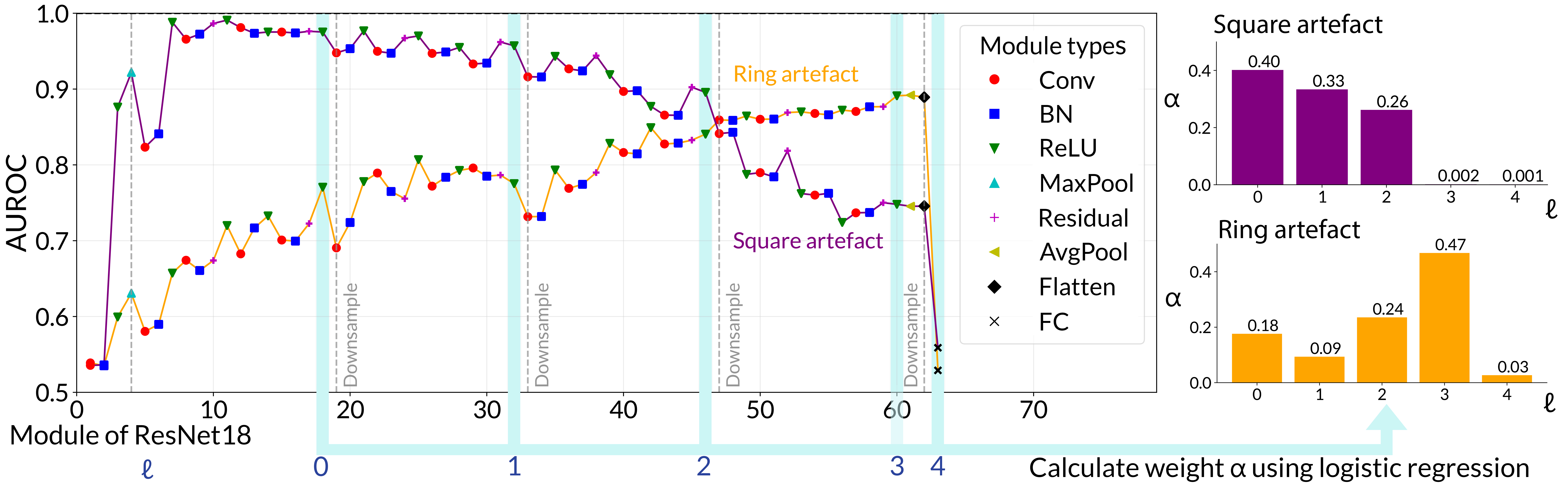} 
\end{tabular}
\caption{AUROC (mean of 3 seeds) for Mahalanobis score over the modules of ResNet18 for synthetic grey square (purple) and white ring (orange) artefacts. The layers used for $\mathcal{D}_{\mathcal{M},\textrm{\textit{comb}}}$ \cite{lee_simple_2018} (Sec. ~\ref{sec:exp_meth}) are highlighted in blue, and the weightings $\alpha_l$ for each layer (Eq.~\ref{coef}) are shown on the right for each artefact. The results show the ideal weighting for one artefact causes a degradation in performance for another - implying there's no one-fits-all weighting.}
\label{square_pacemaker_AUROC}
\end{figure}

\textbf{Ring artefact}: The experiments were repeated with a white ring as the synthetic artefact, and results were compared with the square artefact (Fig. \ref{square_pacemaker_AUROC}). The figure shows the AUROC for different OOD patterns peak at different depths of the network.
The figure shows the layers and optimised linear coefficients $\alpha_l$ for each artefact for $\mathcal{D}_{\mathcal{M},\textrm{\textit{comb}}}$ (Eq.~\ref{coef}), highlighting that the ideal weighting of distances for one OOD pattern can cause a degradation in the performance for another, there is no single weighting that optimally detects both patterns.
As the types of OOD patterns that can be encountered are unpredictable, the idea of searching for an optimal weighting of layers may be ill-advised - implying a different application of this method is required.

\section{Investigation of the use of $\mathcal{D}_{ \mathcal{M}}$ with Real Artefacts}
\label{sec:real_data}

To create an OOD benchmark, we manually labelled 50\% of 
the frontal scans in CheXpert based on whether they had a) no support device, b) any support devices (e.g. central lines, catheters, pacemakers), c) definitely containing a pacemaker, d) unclear. This was performed because CheXpert’s “support devices” class is suboptimal, and to separate pacemakers (distinct OOD pattern).
Findings from the synthetic data were validated on two real OOD tasks (described in Fig.\ref{data_summary}). For the first benchmark, models were trained with scans with no support devices to classify if a scan had Pleural Effusion or not (ID task). Images containing pacemakers were then used as OOD test cases. For the second benchmark, models were trained on males' scans with no support devices to classify for Pleural Effusion, then females' scans with no support devices were used as OOD test cases. For both cases, the datasets were split using 5-fold cross validation, using 80\% of ID images for training and the remaining 20\% as ID test cases.

\begin{figure}
\centering
 \includegraphics[width=\textwidth,height=4.2cm]{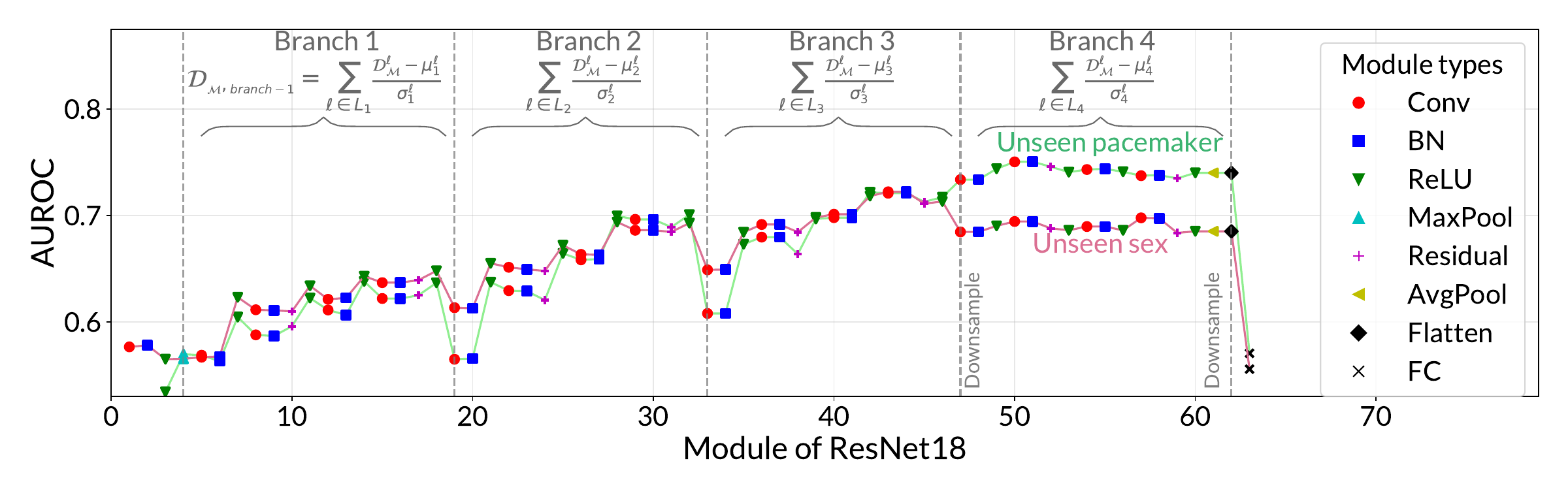} 
\caption{
AUROC (mean of 5 folds) for Mahalanobis score at different modules of ResNet18 for unseen pacemaker (green) and unseen sex (pink) OOD tasks. The figure shows the modules in each branch for MBM with grey brackets.
}
\label{pacemaker_sex_AUROC}
\end{figure}

\textbf{Where to measure $\mathcal{D}_{ \mathcal{M}}$:} Figure \ref{pacemaker_sex_AUROC} shows the AUROC for unseen pacemaker and sex OOD tasks when $\mathcal{D}_{ \mathcal{M}}$ is measured at different modules of a ResNet18. The figure validates the findings on synthetic artefacts: applying $\mathcal{D}_{ \mathcal{M}}$ on the LHL can result in poor performance, and the AUROC performance after a ReLU module is generally improved compared to the preceding BN and convolution. Moreover, it shows that the unseen pacemaker and sex OOD tasks are more detectable at different depths of ResNet18 (modules 51 and 44 respectively). As real-world OOD patterns are very heterogeneous, this motivates an optimal OOD detection system having multiple detectors, each processing features of a network at different layers responsible for identifying different OOD patterns.

\textbf{Compared methods:}
The OOD detection performance of multi-branch Mahalanobis (MBM) was studied. MBM was also investigated using only distances after ReLUs, as experiments on synthetic OOD patterns suggested this may be beneficial. The impact of FGSM (Sec.~\ref{sec:exp_meth}) on MBM was also studied. 
This was compared to OOD detection baselines. The softmax-based methods used were MCP \cite{hendrycks_baseline_2018}, MCDropout \cite{gal_dropout_2016}, Deep Ensembles \cite{lakshminarayanan_simple_2017} (using 3 networks per k-fold),

\begin{table}
\hspace{1.0em}
\caption{AUROC (mean for 5 folds) for OOD detection methods for a) unseen pacemaker and b) unseen sex OOD tasks. \textbf{Bold} highlights the best result of methods, not including oracle methods which represent a theoretical upper bound. * methods with hyperparameters optimised on OOD data.  }
\begin{tabular}{c | c c c c | c c c c }

\multicolumn{9}{c}{a) Unseen pacemaker OOD task} \\
\hline
& \multicolumn{4}{c|}{ResNet18 (AUROC $\uparrow$)} & \multicolumn{4}{c}{VGG16 (AUROC $\uparrow$)} \\
\hline 
\rowcolor{gray!10}
MCP \cite{hendrycks_baseline_2018} &  \multicolumn{4}{c|}{58.4}  & \multicolumn{4}{c}{58.3} \\
Monte Carlo Dropout \cite{gal_dropout_2016}  &  \multicolumn{4}{c|}{58.4}  & \multicolumn{4}{c}{58.4} \\
\rowcolor{gray!10}
Deep Ensemble \cite{lakshminarayanan_simple_2017}  &  \multicolumn{4}{c|}{59.7}  & \multicolumn{4}{c}{60.0} \\
ODIN* \cite{liang_enhancing_2020} &  \multicolumn{4}{c|}{66.1}  & \multicolumn{4}{c}{70.3} \\
\hline 
\rowcolor{gray!10}
Mahal. Score (LHL)\cite{lee_simple_2018}  &  \multicolumn{4}{c|}{57.1}  & \multicolumn{4}{c}{55.8} \\
Mah. Score (LHL) + FGSM\cite{lee_simple_2018}   &  \multicolumn{4}{c|}{57.4}  & \multicolumn{4}{c}{57.5} \\
\rowcolor{gray!10}
Mahal. Score (weight. comb)\cite{lee_simple_2018}  &  \multicolumn{4}{c|}{64.5}  & \multicolumn{4}{c}{66.0} \\
M. Score (w. comb w/o LHL) &  \multicolumn{4}{c|}{71.4}  & \multicolumn{4}{c}{67.4} \\
\hline
\rowcolor{gray!10}
\emph{M. Score (Opt. Layer - Oracle)}* &   \multicolumn{4}{c|}{\emph{75.1} \emph{(after module 51)}}  & \multicolumn{4}{c}{\emph{76.4} \emph{(after module 40)}} \\
\hline
Multi-branch Mahal. (MBM) &   61.9 \ & 66.2 \ & 69.6 \ & 76.1  &  60.4 \ & 60.3 \ & 67.1 \ & 75.0 \\
\rowcolor{gray!10}
MBM (only ReLUs) &   63.6 \ & 68.8 \ & 71.7 \ & 76.2   &  61.2 \ & 63.8 \ & 71.7 \ & 76.2  \\
MBM (only ReLUs) + FGSM* &   63.6 \ & 68.8 \ & 73.1 \ & \textbf{76.8}  & 61.2 \ & 63.8 \ & 74.1 \ & \textbf{77.0}  \\
\hline

\multicolumn{9}{c}{b) Unseen sex OOD task} \\
\hline
& \multicolumn{4}{c|}{ResNet18 (AUROC $\uparrow$)} & \multicolumn{4}{c}{VGG16  (AUROC $\uparrow$)} \\
\hline 
\rowcolor{gray!10}
MCP \cite{hendrycks_baseline_2018} &  \multicolumn{4}{c|}{57.0}  & \multicolumn{4}{c}{56.6} \\
Monte Carlo Dropout \cite{gal_dropout_2016}  &  \multicolumn{4}{c|}{57.0}  & \multicolumn{4}{c}{56.7} \\
\rowcolor{gray!10}
Deep Ensemble \cite{lakshminarayanan_simple_2017}  &  \multicolumn{4}{c|}{58.3}  & \multicolumn{4}{c}{57.7} \\
ODIN* \cite{liang_enhancing_2020} &  \multicolumn{4}{c|}{60.4}  & \multicolumn{4}{c}{64.4} \\
\hline 
\rowcolor{gray!10}
Mahal. Score (LHL) \cite{lee_simple_2018} &  \multicolumn{4}{c|}{55.6}  & \multicolumn{4}{c}{55.2} \\
Mah. Score (LHL) + FGSM\cite{lee_simple_2018} &  \multicolumn{4}{c|}{55.8}  & \multicolumn{4}{c}{57.0} \\
\rowcolor{gray!10}
Mahal. Score (weight. comb)\cite{lee_simple_2018}  &  \multicolumn{4}{c|}{64.3}  & \multicolumn{4}{c}{63.0} \\

M. Score (w. comb w/o LHL)  &  \multicolumn{4}{c|}{70.3}  & \multicolumn{4}{c}{66.7} \\
\hline
\rowcolor{gray!10}
\emph{M. Score (Opt. Layer - Oracle)*}  &  \multicolumn{4}{c|}{\emph{72.2} \emph{(after module 44)}}  & \multicolumn{4}{c}{\emph{76.3} \emph{(after module 43)}} \\
\hline
Multi-branch Mahal. (MBM)  &   63.4 \ & 67.5 \ & 70.8 \ & 70.6  &  62.7 \ & 64.2 \ & 67.8 \ & 74.7  \\
\rowcolor{gray!10}
MBM (only ReLUs) &  64.9 \ & 69.3 \ &  71.8 \ & 70.2  &  63.8  & 66.2 \ & 69.7 \ & 76.4  \\
MBM (only ReLUs) + FGSM* &   64.9 \ & 69.3 \ & \textbf{72.1} \ & 71.4 &  63.8 \ & 66.2 \ & 70.4 \ & \textbf{78.0}  \\
\hline

\end{tabular}
\label{results}
\end{table}

\noindent ODIN \cite{liang_enhancing_2020} (optimising temperature T $\in [1,100]$ and perturbation $\varepsilon \in [0,0.1]$). The performance was also compared to distance-based OOD detection methods such as $\mathcal{D}_{\mathcal{M}}$, $\mathcal{D}_{\mathcal{M , \textrm{\textit{comb}}}}$ ($\alpha_l=1 \ \forall l$), $\mathcal{D}_{\mathcal{M}}$ with FGSM (using an optimised perturbation $\varepsilon \in [0,0.1]$) and $\mathcal{D}_{\mathcal{M}}$ at the best performing network module.

Performance of OOD methods for both ResNet18 and VGG16 are shown in Table \ref{results}. Results show that $\mathcal{D}_{\mathcal{M , \textrm{\textit{comb}}}}$ without LHL outperforms the original weighted combination, showing that the LHL can have a degrading impact on OOD detection. MBM results for ResNet18 in Table \ref{results} show that the OOD patterns are optimally detected at different branches of the network (branch 4 and 3 respectively), further motivating an ideal OOD detector using multiple depths for detecting different patterns. For VGG16 these specific patterns both peak in the deepest branch, but other patterns, such as synthetic squares, peak at different branches (these results are not shown due to space limits). MBM results show that if one could identify the optimal branch for detection of a specific OOD pattern, the MBM approach not only outperforms a sum of all layers, but also outperforms the best performing single layer for a given pattern in some cases. Deducing the best branch for detecting a specific OOD pattern has less degrees-of-freedom than the best layer, meaning an ideal system based on MBM would be easier to configure. The results also show MBM performance can be improved by only using ReLU modules, and optimised with FGSM.

\textbf{Finding thresholds:}
Using multiple OOD detectors poses the challenge of determining OOD detection thresholds for each detector. To demonstrate the potential in the MBM framework, a grid search optimised the thresholds for four OOD detectors of MBM using ReLU modules for ResNet18 trained on setting 3 (described in Fig.\ref{data_summary}). Thresholds were set to classify an image as OOD if any detector labeled it as such. Unseen pacemakers and unseen sex were used as OOD tasks to highlight that thresholds could be found to accommodate multiple OOD tasks. The performance of these combined OOD detectors was compared to $\mathcal{D}_{\mathcal{M , \textrm{\textit{comb}}}}$ w/o LHL ($\alpha_l=1 \ \forall l$) and $\mathcal{D}_{\mathcal{M , \textrm{\textit{comb}}}}$ with optimised $\alpha_l$ (Eq.\ref{coef}) where both require a single threshold, using balanced accuracy as the metric (table \ref{balanced_acc}). Although optimising thresholds for all OOD patterns in complex settings would be challenging,
these results show the theoretically attainable upper bound outperforms both single-layer or weighted combination techniques. Methods for configuring such multi-detector systems can be an avenue for future research.

\begin{table}
\caption{Balanced Accuracy for simultaneous detection of 2 OOD patterns, showing a multi-detector system can improve OOD detection over single-detector systems based on the optimal layer or optimal weighted combination of layers.}
\begin{tabular}{c | c  c  c  }
\hline
\multirow{2}{*}{OOD detection method} & \multicolumn{3}{c}{OOD task (balanced accuracy $\uparrow$)} \\
  & Both tasks & Unseen sex & Pacemakers \\
\hline 
\rowcolor{gray!10}
Mahal. score (equally weighted comb w/o LHL) & 67.64  & 64.63 & 70.37 \\ 
Mahal. score (weighted comb with optimised $\alpha_l$) & 68.14  & 64.89 & 70.90 \\ 
\rowcolor{gray!10}
Multi-branch Mahal. (ReLU only)  & \textbf{71.40} & \textbf{67.26} & \textbf{75.16} \\ 
\hline
\end{tabular}
\label{balanced_acc}
\end{table}

\section{Conclusion}
This paper has demonstrated with both synthetic and real OOD patterns that different OOD patterns are optimally detectable using Mahalanobis score at different depths of a network. The paper shows that the common implementations using the last hidden layer or a weighted combination of layers are sub-optimal, and instead a more robust and high-performing OOD detector can be achieved by using multiple OOD detectors at different depths of the network - informing best-practices for the application of Mahalanobis score. Moreover, it was demonstrated that configuring thresholds for multi-detector systems such as MBM is feasible, motivating future work into developing an ideal OOD detector that encompasses these insights.

\subsubsection{Acknowledgments.} HA is supported by a scholarship via the EPSRC Doctoral Training Partnerships programme [EP/W524311/1]. The authors also acknowledge the use of the University of Oxford Advanced Research Computing (ARC) facility in carrying out this work (http://dx.doi.org/10.5281/zenodo.22558).

%
%
\bibliographystyle{splncs04_with_et_al}
\bibliography{references}

\end{document}